\documentclass[11pt]{article}

\usepackage[preprint]{acl}

\usepackage{times}
\usepackage{latexsym}

\usepackage[T1]{fontenc}

\usepackage[utf8]{inputenc}

\usepackage{microtype}

\usepackage{inconsolata}

\usepackage{graphicx}

\usepackage{booktabs}

\title{SAMPA: Transformer-based segmentation of prosodic boundaries in Brazilian Portuguese}

\author{Rodrigo de Freitas Lima \\
  ICMC \\
  University of São Paulo \\
  São Paulo, Brazil \\
  \texttt{guico21@usp.br} \\\And
  Julio Cesar Galdino \\
  ICMC \\
  University of São Paulo \\
  São Paulo, Brazil \\
  \texttt{juliogaldino@usp.br} \\\And
  Marcos Vinicius Treviso \\
  Instituto Superior Técnico \\
  University of Lisbon \\
  Lisbon, Portugal \\
  \texttt{marcosvtreviso@gmail.com}}

\usepackage{todonotes}

\usepackage{xspace}
\newcommand{\methodname}{SAMPA\xspace}
\newcommand{\fbin}{\ensuremath{F_1^{\mathrm{bin}}}\xspace}
\newcommand{\fmacro}{\ensuremath{F_1^{\mathrm{macro}}}\xspace}

\begin{document}
\maketitle
\begin{abstract}
Automatic prosodic segmentation identifies boundaries between speech units based on acoustic and linguistic evidence. Although recent deep learning approaches have achieved strong results for English, automatic segmentation for Brazilian Portuguese (BP) still relies largely on rule-based or traditional machine-learning methods. This paper presents \methodname, a Whisper-based segmenter that transcribes BP speech while inserting explicit markers for terminal prosodic boundaries. 
We fine-tune Whisper large-v3 on manually segmented recordings from the NURC-Sampa dataset and evaluate different training and test-time filtering configurations, including out-of-distribution testing on the MuPe-Diversidades dataset. Without test-time filtering, the best models reach \fbin of 0.730 on the held-out test split and 0.793 on MuPe-Diversidades, outperforming an out-of-the-box Whisper baseline and a retrained feature-based classifier. Finally, n-gram and acoustic-visual analyses provide qualitative observations that help interpret the model's predictions and formulate hypotheses about the linguistic and prosodic cues that may influence its boundary decisions.
\end{abstract}

\section{Introduction}

Automatic speech segmentation is the task of identifying where boundaries between speech units occur \cite{bigi2018automatic}. When these units are delimited by prosodic cues, they are commonly described as intonation units and play several roles in spoken communication, including structuring discourse and organizing information flow \cite{Raso_Mello_2012}. Automatic segmentation is also of practical interest. Manual annotation of intonation units is time-consuming and requires trained annotators, which limits the size of prosodically segmented corpora. Automatic segmenters can support the annotation of new corpora, research on spoken discourse, and the construction of training data for downstream tasks such as text-to-speech (TTS) synthesis \cite{galdino2025impactprosodicsegmentationspeech}.

Earlier automatic approaches to prosodic segmentation relied on more traditional machine-learning pipelines. For instance, \citet{kocharov17_interspeech} combined acoustic information with syntactic post-processing and reported effective boundary detection for English and Russian. More recently, prosodic segmentation has also benefited from large pretrained speech models. \citet{roll2023psst} introduced PSST!, a Transformer-based approach that fine-tunes Whisper \cite{10.5555/3618408.3619590} to transcribe speech and predict boundary markers within a single output sequence, directly from audio.

For Brazilian Portuguese (BP), \citet{craveiro-etal-2024-simple} adapted heuristic methods originally proposed for English \cite{Biron_etal_2021}, relying mainly on speech rate and silent pauses. Such rules are simple and efficient, but their performance may degrade when parameters are not adapted to the target language or when the audio quality is low.
\citet{Raso_Teixeira_Barbosa_2020} used Linear Discriminant Analysis to identify terminal and non-terminal breaks in BP, while \citet{Craveiro_2025_STIL} proposed a Random Forest classifier based on acoustic features. These studies show that acoustic cues such as fundamental-frequency movement, tessitura changes, and pauses are useful, but they also suggest that some BP intonation units remain difficult to detect with traditional feature-based methods \cite{galdino2026prosodic}.

This gap is particularly relevant in BP, which exhibits substantial regional variation across spoken varieties that may affect prosodic realization. Because prosodic organization is language-specific, methods developed for other languages cannot be assumed to transfer directly to BP \cite{hirst1998survey}, nor can performance on one BP variety be assumed to generalize to others. This motivates the use of large pretrained speech models and the evaluation of their generalization across BP varieties. To the best of our knowledge, no end-to-end model that jointly transcribes speech and predicts prosodic boundaries has been reported for BP. We therefore propose \methodname (Segmenter for Automatic Marking of Prosodic boundAries in BP)\footnote{The acronym SAMPA is also used for the Speech Assessment Methods Phonetic Alphabet, a computer-readable phonetic transcription system. The two uses have different purposes. Here, it was chosen because “Sampa” is a common nickname for São Paulo, the BP variety represented in the test corpus described in Section~\ref{sec:data}.}, a deep learning-based segmenter adapted from PSST! \cite{roll2023psst}. The research question guiding this paper is: how accurately can a Whisper-based model identify terminal intonation-unit boundaries in BP speech? The main contributions of this study are as follows:
\begin{itemize}
    \item The \methodname, a deep learning-based automatic prosodic segmenter for BP adapted from PSST! \cite{roll2023psst}. Unlike prior approaches for BP that rely on acoustic feature extraction or require textual representations and/or forced alignment as input for boundary detection, \methodname predicts both the transcription and prosodic boundaries directly from speech during inference;
    \item A reproducible evaluation protocol that separates evaluation on a held-out test set from a single BP variety from out-of-domain evaluation, allowing us to assess both in-domain performance and cross-dialect generalization across BP varieties, together with comparisons against an out-of-the-box Whisper baseline and a feature-based classifier retrained on the same data;
    \item An exploratory analysis of training-time and test-time acoustic filtering, which shows small differences among configurations that are not consistent across datasets, and complementary analyses of model predictions, combining n-gram analysis of linguistic patterns with targeted acoustic-visual examination of selected cases to investigate patterns associated with boundary predictions.
\end{itemize}

The processed datasets, training and evaluation code, and trained models will be released on the Hugging Face Hub upon acceptance.

\section{Data}
\label{sec:data}

We use two BP datasets with manually annotated prosodic segmentation, both representing the São Paulo variety of BP. The annotations follow the intonation-unit framework adopted in C-ORAL-BRASIL \cite{Raso_Mello_2012}, which distinguishes between terminal boundaries (TB – conclusive) and non-terminal boundaries (NTB – continuative). Since this study focuses on TBs, only terminal boundaries are considered positive instances in our segmentation task. The first dataset is the CORAA NURC-SP Minimal Corpus (MC) \cite{santos22_iberspeech}, which comprises data from the NURC project (\textit{Norma Urbana Linguística Culta}) \cite{oliviera2016nurc}. The original recordings were made on reel-to-reel tapes in the 1970s and subsequently digitized by the TaRSiLA project\footnote{\url{https://sites.google.com/view/tarsila-c4ai}}. Six trained annotators performed and reviewed the segmentation according to the C-ORAL-BRASIL guidelines. Cohen’s kappa values exceeded 0.80 in most cases. The annotated files were subsequently reviewed by a linguistics expert to identify significant deviations from the annotation guidelines.

The second dataset is CATNA (\textit{Corpus de Áudios e Transcrições Não-Alinhadas}) \cite{rodrigues2024portal}, which is publicly maintained by the TaRSiLA project. The corpus underwent automatic prosodic segmentation, which was subsequently reviewed by students, and has the same general configuration as NURC-SP MC\footnote{https://sites.google.com/view/tarsila-c4ai/coraa-versions}. Five recordings, however, were incomplete, while the remaining recordings included an initial audio header that was not annotated in the corresponding Praat TextGrid files. TaRSiLA members provided corrected versions that allowed us to remove the headers and align the audio with the TextGrid annotations. Among these, 11 recordings that had been split into multiple files were reconstructed by merging their parts, and 10 other recordings were trimmed to remove the headers. We refer to this processed dataset as CATNA-MT (Merged and Trimmed).

The collection comprised 47 files: 21 from the CORAA NURC-SP MC and 26 from CATNA-MT. Five incomplete CATNA recordings were excluded. The CATNA recording SP\_D2\_012\_1 was excluded because it was judged perceptually unintelligible upon listening. The CORAA NURC-SP MC recording SP\_D2\_062 was excluded because its metadata indicated poor audio quality. The combination of the two datasets is referred to as NURC-Sampa throughout this article (Table \ref{tab:characteristics-dataset}).

\section{Model}
\label{sec:model}

\methodname is obtained by fine-tuning the Whisper Large-v3 model \cite{10.5555/3618408.3619590} for the task of prosodic boundary prediction in BP, following the methodology proposed by \citet{roll2023psst} for English. This section describes the base model and the adaptation strategy adopted in this work. Whisper \cite{10.5555/3618408.3619590} is an automatic speech recognition (ASR) system based on the Transformer encoder-decoder architecture \cite{10.5555/3295222.3295349}, trained in a weakly supervised fashion on a large multilingual corpus of 680,000 hours of audio collected from the internet. This scale of training confers high robustness to different acoustic conditions and accents, as well as competence across dozens of languages, including BP. In this work, we use the \textit{large-v3} variant,\footnote{\url{https://huggingface.co/openai/whisper-large-v3}} the largest multilingual Whisper checkpoint available at the time of our experiments.

\begin{figure}[htbp]
\centering
\includegraphics[width=\columnwidth]{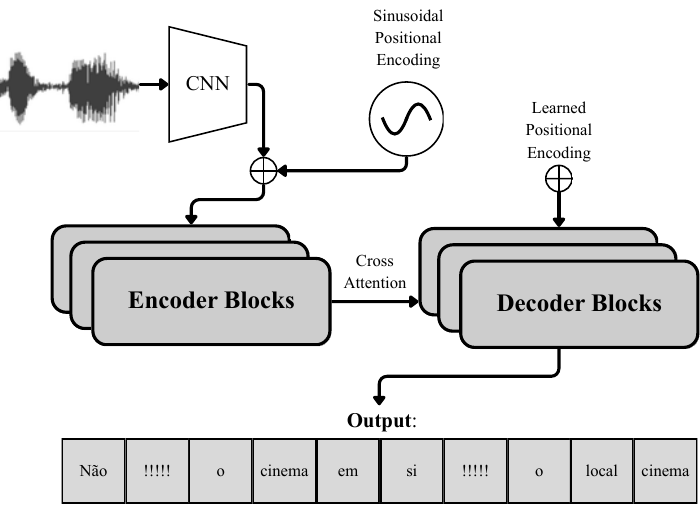}
\caption{Overview of \methodname. The input waveform is converted into log-Mel features, encoded by the Whisper encoder, and decoded as a transcription containing both words and the delimiter token used to mark predicted terminal prosodic boundaries.}
\label{fig:architecture}
\end{figure}

The key idea behind \methodname is to cast prosodic segmentation as an extension of speech transcription. Instead of producing only lexical text, the model is trained to output a transcription that includes explicit boundary markers. Thus, transcription and prosodic segmentation are learned within a single sequence-to-sequence architecture.
During fine-tuning, terminal boundary positions are represented by a special delimiter token (``!!!!!'' in our case) inserted between transcribed words. For example, an audio clip containing two consecutive intonation units is represented as \textit{``first unit !!!!! second unit''}. 
Importantly, no architectural changes to Whisper are required: the delimiter is added to the vocabulary as a decoder output token, allowing boundary predictions to be conditioned on the encoded audio and the preceding output sequence. Figure~\ref{fig:architecture} illustrates the resulting architecture.

\section{Experimental Details}
\label{sec:preprocessing-and-training}

To make the original datasets compatible with the PSST! training procedure, we first parsed the annotation files to extract the start time, end time, speaker, and text of each annotated segment. To generate the training samples, consecutive segments of the same speaker were concatenated while the resulting audio remained below Whisper's 30-second input limit, since longer inputs are truncated and boundaries near the end of the sample would be lost.

\subsection{Preprocessing pipeline}

Adjacent segments were grouped according to three conditions: (i) preservation of the original file ordering, (ii) speaker continuity, and (iii) the maximum duration constraint. When all conditions were satisfied, the texts of the segments were concatenated using the delimiter token ``!!!!!''; otherwise, a new sample was created. Since the original audios were recorded on tape, a portion of the recordings contained a significant amount of noise, making the derived samples unsuitable for training. Therefore, the recordings were manually checked, and a recording was considered too noisy when it was impossible for the researchers to understand the words being said.

Four different recordings ($\sim9\%$ of the original data, $\sim3$ hours of speech) were held out to generate the test split. The recordings were chosen so that there would be two audios in which the main speaker was male and two in which it was female, with the aim to minimize gender bias. Furthermore, each gender has one clean recording and one with moderate noise, so that the model could be evaluated on both clean and noisy recordings. The identifiers of the test recordings are listed in Appendix~\ref{sec:appendix-details}. The rest of the data ($\sim29$ hours) was used to generate the train/validation split. 
The resulting dataset was converted into the format required for Whisper-based training. Audio signals were resampled to 16~kHz and paired with the corresponding transcriptions containing the prosodic boundary markers.

For out-of-distribution evaluation, we use MuPe-Diversidades \cite{Galdino2025Mupe}, which is also prosodically segmented. This dataset contains 2h~42m of audio and differs from the NURC-Sampa data in two important ways: its recordings are substantially cleaner, and its speakers are more geographically diverse. Whereas most NURC-Sampa speakers were born in the state of S\~ao Paulo and therefore tend to represent the São Paulo variety, MuPe-Diversidades includes samples from 17 Brazilian states. Table \ref{tab:characteristics-dataset} describes the characteristics of our datasets after preprocessing.

\begin{table}[t]
\caption{Characteristics of our datasets. \textit{Segments} are the annotated units after preprocessing, not terminal boundaries.}
\label{tab:characteristics-dataset}
\centering
\resizebox{\columnwidth}{!}{%
\begin{tabular}{llrrrr}
\toprule
\textbf{Dataset} & \textbf{Split} & \textbf{Recordings} & 
\textbf{Segments} & \textbf{Duration} & \textbf{Seg/min} \\
\midrule
NURC-Sampa & Train/Validation & 36 & 11,878 & 28h 44m & $\approx$6.9 \\
NURC-Sampa & Test & 4 & 789 & 2h 58m & $\approx$4.4 \\
MuPe-Div. & Full & 30 & 1,099 & 2h 42m & $\approx$6.8 \\
\bottomrule
\end{tabular}%
}
\end{table}

\subsection{Training}

Because several recordings in the training data contain noise, we also created alternative dataset versions by applying digital filters. Low-pass (LP) and high-pass (HP) filters were applied with different cutoff frequencies, yielding five training configurations: unfiltered audio, LP 3200~Hz, HP 400~Hz, HP 600~Hz, and data augmentation combining the four previous versions.

The cutoff frequencies were selected from preliminary experiments with a model trained on unfiltered data, where these settings produced the highest binary $F_1$ values on filtered versions of the test audio. Because this selection used the held-out test audio, the comparison between filtering configurations is not independent of the test set. We therefore report the unfiltered test condition as our primary result and treat the filtering results as an exploratory analysis (Section~\ref{sec:filters}). At inference time, each model was also evaluated on versions of the test audio processed with LP filters (200, 400, 800, 1,600, and 3,200~Hz cutoffs) and HP filters (200, 400, 600, 800, 1,200, 1,600, and 3,200~Hz cutoffs). All filters were implemented as fourth-order Butterworth filters (SciPy), applied to the audio resampled to 16~kHz.
We trained five independent models, one for each configuration above. From the training data, 5\% of the samples were randomly held out for validation, and samples shorter than 1 second were discarded. In the data augmentation configuration, the four versions of each training sample were concatenated, quadrupling the size of the training set, whereas the validation and test sets always used the original, unfiltered audio. All models were trained for 4 epochs. Following PSST!, the learning rate was increased linearly to $10^{-5}$ during the initial warm-up steps (about 7\% of the total steps) to reduce early overfitting and then decayed linearly. The model was evaluated on the validation set at every 12.5\% of the training steps, and the checkpoint with the highest \fbin on the validation set was selected. Further implementation details, including batch size, optimizer, hardware, and decoding parameters, are given in Appendix~\ref{sec:appendix-details}.

\subsection{Baselines}
\label{sec:baselines}

We compare \methodname with three baselines. The \textit{no-boundary} baseline never predicts a boundary and indicates the scores obtained from the majority class alone. The \textit{out-of-the-box Whisper} baseline uses the original Whisper large-v3 model without fine-tuning, following PSST!; we treat periods, exclamation marks and question marks in its output as predicted boundaries. Since Whisper already produces sentence-final punctuation, this baseline indicates how much of the segmentation performance can be obtained without task-specific fine-tuning. Finally, we compare \methodname with ProsSegue-ML \cite{Craveiro_2025_STIL}, a Random Forest classifier based on acoustic features that requires a reference transcription as input, retrained on the NURC-Sampa training split.

\section{Results and Discussion}

Prosodic boundary detection is evaluated as a binary classification problem: for each potential boundary position between corresponding words, the model must decide whether a terminal prosodic boundary is present. Following \citet{galdino2026prosodic}, we compare each generated token sequence with its reference by checking the presence or absence of the boundary marker between aligned words. Since the generated transcription may differ from the reference, both sequences are tokenized by whitespace, with the boundary marker treated as a separate token, and aligned with the \texttt{ndiff} function of Python's \texttt{difflib} library. A marker present in both aligned sequences is counted as a true positive, a marker present only in the reference as a false negative, a marker present only in the prediction as a false positive, and each word matched in both sequences as a true negative. Words inserted, deleted, or substituted by the model are ignored, since they do not correspond to boundary positions. An example is given in Appendix~\ref{sec:appendix-alignment}. We report Precision (P), Recall (R), binary F1 (\fbin), and macro F1 (\fmacro). The \fbin score is computed over the positive class, namely boundary positions. The \fmacro score averages the F1 scores of the boundary and non-boundary classes, penalizing both excessive insertion and omission of boundaries. We adopt \fbin as our primary metric because terminal boundaries correspond to a small fraction of the candidate positions, so the non-boundary class inflates \fmacro: a system that never predicts a boundary already obtains \fmacro of approximately 0.49 (Tables~\ref{tab:results-test} and~\ref{tab:results-mupe}). We also report Word Error Rate (WER), computed after removing the boundary markers, to monitor whether segmentation fine-tuning harms transcription quality. Unless otherwise stated, all results are obtained on unfiltered test audio. To assess whether the differences between configurations are statistically significant, we use a paired bootstrap test, resampling the audio samples of each evaluation set (672 in the held-out test split and 906 in MuPe-Diversidades, 10,000 resamples) and computing the difference in \fbin between configurations on the same resamples.

\subsection{In-domain evaluation}

Table~\ref{tab:results-test} presents the results on the held-out test split without test-time filtering. All \methodname configurations outperform the three baselines by a large margin: the best configuration, LP 3200~Hz, obtains \fbin of 0.730 and \fmacro of 0.857, compared with 0.479 and 0.723 for the out-of-the-box Whisper and 0.362 and 0.661 for the retrained ProsSegue-ML. Compared with the out-of-the-box Whisper, \methodname improves both precision (0.612--0.638 vs. 0.372) and recall (0.812--0.879 vs. 0.671). In all configurations, recall is higher than precision, indicating that most errors are boundary insertions rather than omissions.
In contrast, the differences among training configurations are small, with \fbin between 0.709 and 0.730. According to the paired bootstrap test, the LP 3200~Hz and HP 600~Hz configurations outperform the unfiltered training configuration ($p=0.004$ and $p=0.016$, respectively, without correction for multiple comparisons), whereas the differences among the four configurations trained on filtered audio are not significant ($p>0.2$).
These results indicate that, for the held-out data, training on filtered audio yields a small advantage over training on the original noisy data. However, since the training cutoffs were selected on filtered versions of the test audio, this advantage may be overestimated.

\begin{table}[t]
\caption{Results on the held-out test split of NURC-Sampa without test-time filtering. ProsSegue-ML uses the reference transcription as input; therefore, no WER is reported.}
\label{tab:results-test}
\centering
\resizebox{\columnwidth}{!}{%
\begin{tabular}{lccccc}
\toprule
\textbf{Model} & \textbf{WER} & \textbf{P} & \textbf{R} & \textbf{\fbin} & \textbf{\fmacro} \\
\midrule
No boundary & -- & 0.000 & 0.000 & 0.000 & 0.488 \\
Whisper (no fine-tuning) & 0.133 & 0.372 & 0.671 & 0.479 & 0.723 \\
ProsSegue-ML (retrained) & -- & 0.271 & 0.542 & 0.362 & 0.661 \\
\midrule
\multicolumn{6}{l}{\textit{\methodname, training configuration:}} \\
Unfiltered & 0.104 & 0.629 & 0.812 & 0.709 & 0.846 \\
LP 3200~Hz & 0.103 & 0.638 & 0.852 & 0.730 & 0.857 \\
HP 400~Hz & 0.107 & 0.612 & 0.879 & 0.722 & 0.853 \\
HP 600~Hz & 0.105 & 0.630 & 0.851 & 0.724 & 0.854 \\
Data aug. & 0.107 & 0.638 & 0.831 & 0.722 & 0.853 \\
\bottomrule
\end{tabular}%
}
\end{table}

Table~\ref{tab:results-noise} reports \fbin separately for each recording of the held-out test split. In all configurations, the clean recordings obtained lower \fbin than the noisy ones (0.643--0.665 vs. 0.758--0.796). However, this difference is mainly due to one clean recording, SP\_DID\_161, for which \fbin ranges from 0.408 to 0.436, whereas the other clean recording, SP\_DID\_234, obtains the highest scores of the test split (0.810--0.840). Thus, audio quality alone does not explain the differences between recordings, and recording-specific factors, such as speaker and speaking style, may have a stronger influence. Since each condition contains only two recordings, this comparison should be interpreted with caution.

\begin{table}[t]
\caption{\fbin on each recording of the held-out test split, without test-time filtering. Recordings are identified by the number in their SP\_DID identifiers (Appendix~\ref{sec:appendix-details}), and the \textit{All} columns aggregate the two recordings of each condition.}
\label{tab:results-noise}
\centering
\resizebox{\columnwidth}{!}{%
\begin{tabular}{lcccccc}
\toprule
 & \multicolumn{3}{c}{\textbf{Clean}} & \multicolumn{3}{c}{\textbf{Noisy}} \\
\cmidrule(lr){2-4} \cmidrule(lr){5-7}
\textbf{Training} & \textbf{234} & \textbf{161} & \textbf{All} & \textbf{011} & \textbf{090} & \textbf{All} \\
\midrule
Unfiltered & 0.810 & 0.437 & 0.654 & 0.769 & 0.730 & 0.758 \\
LP 3200~Hz & 0.823 & 0.426 & 0.652 & 0.810 & 0.760 & 0.796 \\
HP 400~Hz & 0.828 & 0.408 & 0.643 & 0.811 & 0.739 & 0.792 \\
HP 600~Hz & 0.831 & 0.434 & 0.661 & 0.799 & 0.727 & 0.780 \\
Data aug. & 0.840 & 0.436 & 0.665 & 0.796 & 0.699 & 0.770 \\
\bottomrule
\end{tabular}%
}
\end{table}

\subsection{Out-of-domain evaluation}

Table~\ref{tab:results-mupe} reports the results on MuPe-Diversidades without test-time filtering. As in the in-domain evaluation, all \methodname configurations outperform the baselines, with \fbin between 0.766 and 0.793, compared with 0.491 for the out-of-the-box Whisper and 0.351 for the retrained ProsSegue-ML. Unlike in the held-out test split, precision is higher than recall in all configurations (0.845--0.864 vs. 0.700--0.736), indicating that boundary omissions are more frequent than insertions on this dataset. The ranking of the training configurations also differs from the in-domain evaluation. The HP 600~Hz configuration obtains the highest \fbin (0.793) and outperforms the unfiltered ($p=0.013$), LP 3200~Hz ($p<0.001$), and data augmentation ($p=0.005$) configurations, whereas the LP 3200~Hz configuration, which obtained the highest score on the held-out test split, obtains the lowest score on MuPe-Diversidades (0.766). Thus, no training configuration is consistently superior across the two datasets, although the HP 600~Hz configuration is among the best configurations in both evaluations.

\begin{table}[t]
\caption{Results on MuPe-Diversidades without test-time filtering. ProsSegue-ML uses the reference transcription as input; therefore, no WER is reported.}
\label{tab:results-mupe}
\centering
\resizebox{\columnwidth}{!}{%
\begin{tabular}{lccccc}
\toprule
\textbf{Model} & \textbf{WER} & \textbf{P} & \textbf{R} & \textbf{\fbin} & \textbf{\fmacro} \\
\midrule
No boundary & -- & 0.000 & 0.000 & 0.000 & 0.481 \\
Whisper (no fine-tuning) & 0.165 & 0.453 & 0.536 & 0.491 & 0.724 \\
ProsSegue-ML (retrained) & -- & 0.421 & 0.301 & 0.351 & 0.658 \\
\midrule
\multicolumn{6}{l}{\textit{\methodname, training configuration:}} \\
Unfiltered & 0.125 & 0.856 & 0.717 & 0.780 & 0.882 \\
LP 3200~Hz & 0.134 & 0.846 & 0.700 & 0.766 & 0.875 \\
HP 400~Hz & 0.130 & 0.845 & 0.736 & 0.787 & 0.886 \\
HP 600~Hz & 0.125 & 0.864 & 0.733 & 0.793 & 0.889 \\
Data aug. & 0.135 & 0.860 & 0.705 & 0.775 & 0.879 \\
\bottomrule
\end{tabular}
}
\end{table}

On MuPe-Diversidades, the best \methodname configuration (HP 600~Hz) achieves \fbin of 0.793 and \fmacro of 0.889 without test-time filtering. In comparison, \citet{Craveiro_2025_STIL} reported binary and macro F1 scores of 0.55 and 0.77, respectively, for ProsSegue-ML on the same corpus. Importantly, the experimental settings differ: \methodname is evaluated without task-specific training on MuPe-Diversidades and predicts both transcription and prosodic boundaries directly from audio, whereas ProsSegue-ML is trained on 80\% of the MuPe corpus and requires a reference transcription. When retrained on our training data and evaluated under the same protocol as \methodname, ProsSegue-ML obtained \fbin of 0.351 on MuPe-Diversidades (Table~\ref{tab:results-mupe}). The difference from the originally reported score is likely related to the absence of MuPe-Diversidades data in training. Thus, \methodname achieves higher scores without using training data from the out-of-domain evaluation corpus or requiring reference transcriptions at inference time. These results suggest that a Transformer-based sequence-to-sequence approach can be effective for prosodic segmentation in BP. A further point of comparison is the robustness evaluation presented by \citet{craveiro-aluisio-2026-robustness}, which highlights the importance of assessing model performance across different corpora. The MuDi model introduced in that study achieved a macro F1 score of 0.75 on MuPe-Diversidades, 0.71 on samples from C-ORAL-BRASIL I and II, and 0.61 on the NURC-SP MC. \methodname, in turn, achieved \fmacro of 0.857 on the held-out test set of NURC-Sampa and 0.889 on MuPe-Diversidades. Although these results provide complementary evidence of \methodname's performance across evaluation datasets, direct comparisons with MuDi should be made with caution because the reported scores may reflect differences in experimental settings and evaluation protocols. Unlike what would be expected for out-of-distribution data, \methodname obtains higher scores on MuPe-Diversidades than on the held-out test split. One possible factor is the density of prosodic boundaries. MuPe-Diversidades contains 1,710 terminal boundaries in 2h~42m of audio (approximately 10.6 per minute), whereas the held-out test split contains 998 terminal boundaries in 2h~58m (approximately 5.6 per minute). Since \fbin depends on the proportion of positive instances, the higher proportion of boundary positions in MuPe-Diversidades (approximately 7.5\% vs. 4.5\% of the candidate positions) may contribute to its higher scores. Another possible factor is audio quality: the NURC recordings originate from reel-to-reel tapes, whereas the MuPe-Diversidades recordings are more recent and have better acoustic quality. However, within the held-out test split, the clean recordings did not obtain higher scores than the noisy ones (Table~\ref{tab:results-noise}), which suggests that audio quality alone does not explain this difference. These factors therefore remain hypotheses.

\subsection{Exploratory analysis of acoustic filtering}
\label{sec:filters}

\begin{figure}[t]
\centering
\includegraphics[width=0.9\columnwidth]{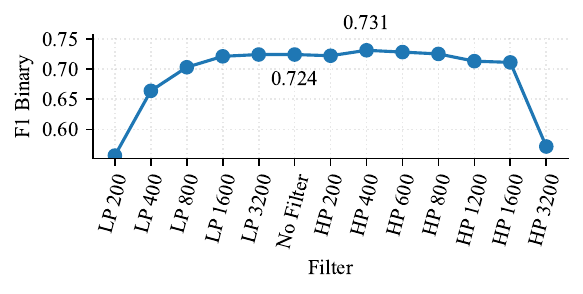}
\includegraphics[width=0.9\columnwidth]{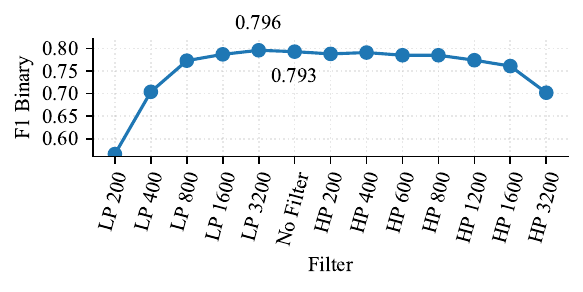}
\caption{\fbin for each test-time filter configuration for the HP 600~Hz model on the test split of NURC-Sampa (top) and on MuPe-Diversidades (bottom). The values for the unfiltered condition (no filter) and the highest \fbin achieved are highlighted. 
}
\label{fig:test_filters}
\end{figure}

Since the filter cutoffs were selected on the held-out test audio (Section~\ref{sec:preprocessing-and-training}), the results in this section should be interpreted as exploratory. On the held-out test split, for the four models trained on filtered audio, the best test-time filter improved \fbin over the unfiltered test condition by 0.1 to 0.8 percentage points, and none of these gains was significant according to the paired bootstrap test ($p>0.16$). For instance, for the HP 600~Hz model, the best filtered condition and the unfiltered condition differed by approximately 0.6 percentage points (0.731 vs. 0.724; Figure~\ref{fig:test_filters}, top). The largest gain, 2.1 percentage points, was obtained for the model trained on unfiltered audio with an HP 600~Hz filter ($p<0.001$). However, the cutoffs were selected with a model trained on unfiltered data, so this gain is directly affected by the selection on the test audio. On MuPe-Diversidades, the best test-time filter improved \fbin by up to 1.3 percentage points. For the HP 600~Hz model, the unfiltered condition remained close to the best condition, with a difference of about 0.2 percentage points (Figure~\ref{fig:test_filters}, bottom). The gains were significant for the unfiltered (1.2 percentage points, $p=0.017$) and HP 400~Hz (0.9 percentage points, $p=0.039$) models, but since the best filter was selected among 12 filters on the evaluation data itself, these $p$-values are optimistic. Overall, these results suggest that test-time filtering yields at most small gains, which are not consistent across models and datasets. The best test-time filter for each training configuration, selected by \fbin, is reported in Appendix~\ref{sec:appendix-filters}.

\subsection{N-gram analysis}

To better understand model behavior, we performed an n-gram analysis of the linguistic contexts in which \methodname marked boundaries. We compared the training configurations by examining correct boundary detections (true positives, TPs) and incorrectly inserted boundaries (false positives, FPs). The distributions were similar across the unfiltered, high-pass, low-pass, and data-augmentation configurations (Table \ref{tab:ngrams}). For TPs, unigrams immediately before predicted boundaries included discourse markers, such as \textit{né}, and frequent nouns such as \textit{cinema}, \textit{filme}, and \textit{casa}; unigrams immediately after TPs included conjunctions such as \textit{e}, \textit{mas}, and \textit{porque}. Given the low n-gram counts, this analysis is qualitative and seeks to formulate hypotheses regarding the linguistic factors that may explain the model's predictions. In the case of unigrams, the position of certain elements, such as discourse markers, may be associated with differences in their prosodic realization \cite{raso2014prosodic}. The presence of nouns before true boundaries (TPs) and connectives after them may also suggest the use of morphosyntactic cues, given that connectives tend to be function words \cite{lyons1981language}. Regarding the trigram analysis, the presence of subject-verb constructions suggests that syntactic information may have influenced \methodname's predictions. Although connectives also appear after false boundaries (FPs), the examples do not reveal improbable word sequences at the predicted boundaries, such as articles at the end of a boundary or unlikely combinations at the start of the following segment. This observation is consistent with the hypothesis that the model's predictions are influenced by linguistic context, even though the low counts do not allow for a definitive confirmation of this hypothesis.

\begin{table}[t]
\centering
\caption{Most frequent unigrams and trigrams before and after true-positive (TP) and false-positive (FP) prosodic boundaries generated by the HP 600~Hz \methodname{} model. We report full counts in parenthesis.}
\label{tab:ngrams}
\scriptsize
\setlength{\tabcolsep}{2pt}
\resizebox{\columnwidth}{!}{%
\begin{tabular}{llll}
\toprule
\multicolumn{2}{c}{\textbf{Before}} & \multicolumn{2}{c}{\textbf{After}} \\
\cmidrule(lr){1-2} \cmidrule(lr){3-4}
\textbf{TP} & \textbf{FP} & \textbf{TP} & \textbf{FP} \\
\midrule
\multicolumn{4}{l}{\textit{\textcolor{gray}{Unigrams:}}} \\
né (82)     & não (25)     & e (119)     & e (59)      \\
não (27)    & né (18)      & eu (91)     & eu (34)     \\
é (24)      & é (11)       & então (81)  & mas (29)    \\
teatro (15) & cinema (10)  & mas (78)    & então (28)  \\
cinema (14) & sucesso (10) & não (52)    & não (28)    \\
de (13)     & também (8)   & é (50)      & o (27)      \\
filme (13)  & filme (8)    & agora (46)  & porque (22) \\
muito (12)  & entende (7)  & o (37)      & é (18)      \\
mesmo (12)  & nada (6)     & porque (34) & ah (14)     \\
casa (11)   & teatro (6)   & quer (33)   & ele (14)    \\
\midrule
\multicolumn{4}{l}{\textit{\textcolor{gray}{Trigrams:}}} \\
lá em casa (3)     & em são paulo (4)          & eu acho que (16) & eu acho que (9)    \\
eu não sei (2)     & no teatro não (2)         & eu não sei (7)   & acho que todo (2)  \\
a mesma coisa (2)  & muito pelo contrário (2)  & como é que (5)   & que que você (2)   \\
de todo mundo (2)  & é bem aceito (2)          & quer dizer o (4) & eu tenho a (2)     \\
mais ou menos (2)  & gostou do filme (2)       & eu tenho ido (4) & o teatro é (2)     \\
\bottomrule
\end{tabular}
}
\end{table}

\subsection{Prosodic analysis}

\begin{figure}[t]
\centering
\includegraphics[width=\columnwidth]{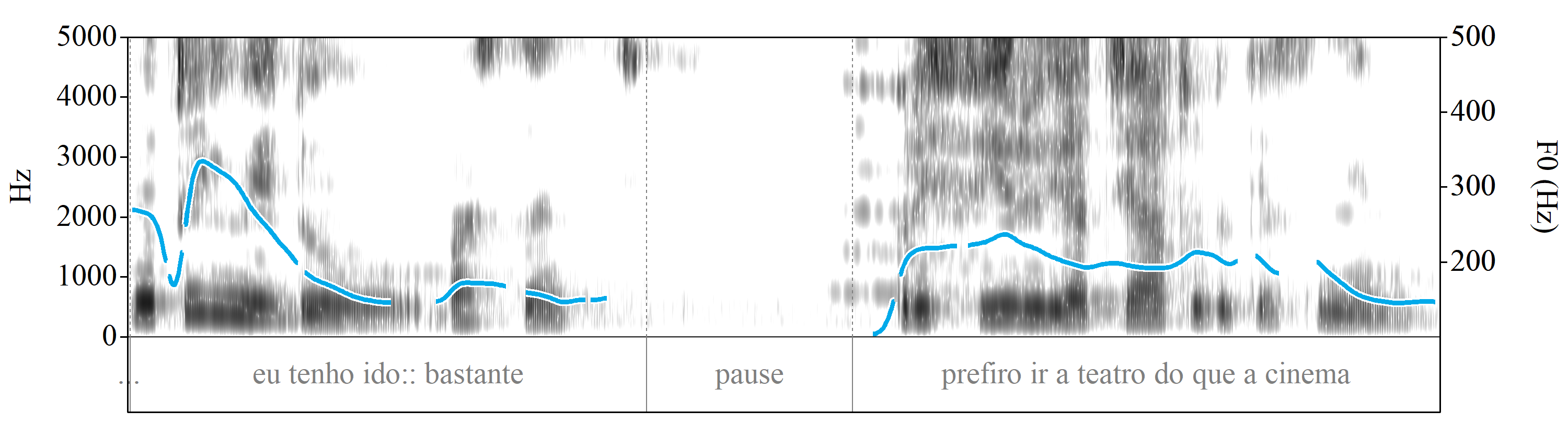}\\[6pt]
\includegraphics[width=\columnwidth]{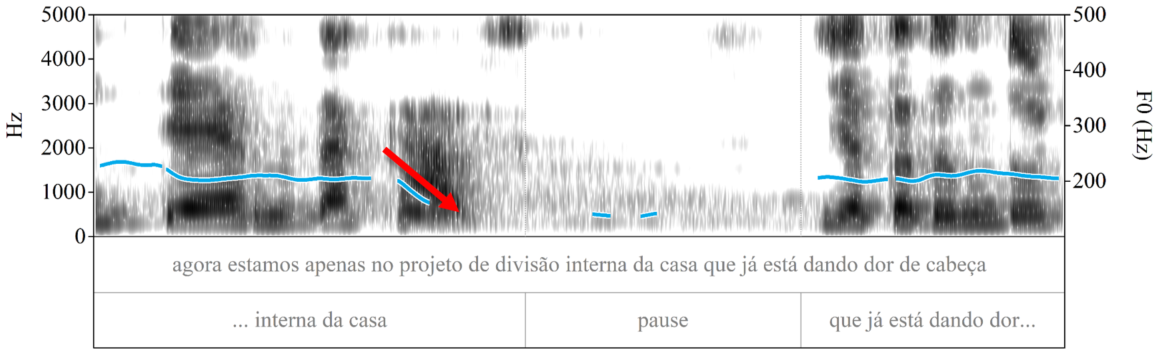}
\caption{Example of correct markings of two different intonation units (top), and a false positive case (bottom).}
\label{fig:positivev}
\end{figure}

Finally, we conducted an acoustic-visual analysis of correct and incorrect segmentations to qualitatively inspect the boundaries predicted by \methodname. We examined test recordings in Praat \cite{boersma2011praat} and analyzed the cases in which the model inserted a boundary between intonation units. Figure~\ref{fig:positivev} illustrates a true positive example (top), where several prosodic cues support the reference boundary, along with a false positive example (bottom), where \methodname inserts a boundary that is not present in the reference annotation. In the true positive example, the F0 contour shows a continuous fall at the end of the first terminal unit (\textit{eu tenho ido bastante}), a pattern typical of BP declaratives \cite{cunha2000entoacao,tenani2002dominios,frota2015intonational}. The pitch range also changes between the two units: the first ends in a lower range, and the following unit begins after a reset to a higher region. These cues are accompanied by a long silent pause. Together, they provide strong evidence for a terminal prosodic boundary. In contrast, in the false positive example, the F0 curve falls on \textit{casa} (marked by the arrow), but this local contour alone is not enough to characterize a terminal boundary. Differences in F0 before and after a candidate boundary can be relevant, but they are not sufficient to determine the boundary on their own \cite{teixeira2018correlatos}. Human annotation also involves perceptual judgment and the informational value of the unit \cite{santos2022hipersegmentaccao}. More generally, intonation-unit identification is difficult because several phonetic-acoustic cues may contribute to boundary perception, and there is no full consensus on which cues distinguish different boundary types or how they interact \cite{teixeira2018correlatos}.

A pause is also present in this example, but pauses may occur at both terminal and non-terminal boundaries \cite{teixeira2025silentpause}. They may create discourse effects such as expectation or emphasis and are neither sufficient nor always necessary for boundary marking \cite{mendes2015papel}. Thus, pauses are important cues, but they can also lead a model to confuse boundary types \cite{Raso_Teixeira_Barbosa_2020}. In addition, the segment \textit{que já está dando uma dor de cabeça} can be interpreted as an explanatory relative clause and could form an autonomous unit in some contexts. In this particular case, however, the annotators did not judge the prosodic and discourse evidence as sufficient to mark two distinct terminal units. Overall, the examined cases suggest that some false positive predictions coincide with plausible acoustic and linguistic cues rather than occurring at entirely arbitrary positions. In particular, some predicted boundaries exhibit acoustic characteristics that may be associated with prosodic boundaries, even when these cues are not sufficient to establish terminal intonation-unit boundaries according to the reference annotation.
This observation is consistent with the annotation framework, which considers not only acoustic evidence but also syntactic and pragmatic aspects of speech. A major challenge for future automatic prosodic segmentation is therefore to incorporate syntactic and discourse-level information more explicitly.

\section{Conclusion}

This paper presented \methodname, a Whisper-based prosodic segmenter for BP adapted from PSST!, which predicts transcription and terminal prosodic boundaries directly from speech. Without test-time filtering, the best configurations achieved \fbin of 0.730 on the held-out NURC-Sampa test split and 0.793 on MuPe-Diversidades, outperforming an out-of-the-box Whisper baseline and a retrained ProsSegue-ML classifier. Training- and test-time filtering yielded small differences that were not consistent across datasets. Finally, the n-gram and acoustic-visual analyses illustrated the difficulty of distinguishing terminal boundaries from weaker prosodic divisions when acoustic, syntactic, and pragmatic cues are ambiguous.

\section*{Limitations}

The main limitation of this study concerns the comparison of filtering configurations. The filter cutoffs were selected on filtered versions of the held-out test audio, and each configuration was trained with a single random seed; to mitigate this, we report the unfiltered test condition as our primary result, treat the filtering results as exploratory, and assess the differences with a paired bootstrap test, which nevertheless does not capture variation across training runs. The held-out test split contains only four recordings, which we addressed by reporting results per recording and by evaluating on an out-of-domain corpus. The filters are not substitutes for neural speech enhancement or for other augmentation techniques, such as time stretching or pitch shifting, which were not evaluated, and we did not perform the lexical and masked-audio controls proposed in PSST!. Since the original ProsSegue-ML model could not be used, we retrained it on our data and evaluated it under the same protocol as \methodname, although it still requires a reference transcription. Finally, the prosodic analysis is limited to selected cases, so we present its observations as hypotheses rather than as quantitative evidence.

\section*{AI Disclosure}

We used AI tools, specifically ChatGPT 5.5 Pro and Claude Opus 5.5, during the writing of this paper for English checking and polishing (grammar enhancement). We also used GitHub Copilot and Claude Code to assist with coding.



\bibliography{custom}

\appendix
\clearpage 
\newpage

\section{Implementation Details}
\label{sec:appendix-details}

All models were fine-tuned with the Hugging Face Transformers library, using a batch size of 4 with 2 gradient accumulation steps (effective batch size of 8), the AdamW optimizer, a peak learning rate of $10^{-5}$ with linear warm-up (7\% of the total steps, with a minimum of 50 steps) followed by linear decay, gradient clipping at a norm of 1.0, and 32-bit floating-point precision. Training was performed on a single NVIDIA RTX A6000. All audio samples were zero-padded to 30 seconds, and the decoder was constrained to Portuguese transcription. Whisper large-v3 has approximately 1.55B parameters, and each model took approximately 9 hours to train. At inference, we used greedy decoding (beam size of 1), with a maximum output length of 225 tokens. Table~\ref{tab:test-recordings} lists the recordings of the held-out test split.

\begin{table}[h]
\caption{Recordings in the held-out test split of NURC-Sampa.}
\label{tab:test-recordings}
\centering
\resizebox{\columnwidth}{!}{%
\begin{tabular}{llll}
\toprule
\textbf{Recording} & \textbf{Corpus} & \textbf{Main speaker} & \textbf{Audio} \\
\midrule
SP\_DID\_234 & NURC-SP MC & Female & Clean \\
SP\_DID\_161 & NURC-SP MC & Male & Clean \\
SP\_DID\_011 & CATNA-MT & Male & Moderate noise \\
SP\_DID\_090 & CATNA-MT & Female & Moderate noise \\
\bottomrule
\end{tabular}%
}
\end{table}

\section{Boundary Alignment Example}
\label{sec:appendix-alignment}
 
The following example is taken from the held-out test split (recording SP\_DID\_090), with the prediction generated by the model trained on unfiltered audio, without test-time filtering:
 
\begin{itemize}
    \item Reference: \textit{fora éh no tem uma parte do quintal que o piso é de aquela pedra mineira !!!!! nessa parte que entra o carro !!!!! agora nos fundos é cimento cimento comum}
    \item Prediction: \textit{ah fora eh tem uma parte do quintal que o piso é de aquela pedra mineira !!!!! nessa parte que entra o carro agora nos fundos é cimento !!!!! cimento comum}
\end{itemize}
 
After alignment with \texttt{ndiff}, the 27 words matched in both sequences are counted as true negatives (TN~=~27). The differences in the transcription at the beginning of the sample (\textit{ah}, \textit{eh}, \textit{éh}, and \textit{no}) appear as insertions and deletions and are ignored. The marker after \textit{mineira} is matched in both sequences (TP~=~1). The marker after \textit{carro} appears only in the reference (FN~=~1), whereas the marker predicted between the two occurrences of \textit{cimento} appears only in the prediction (FP~=~1). Thus, a boundary placed at a different position from the reference is counted as one false negative and one false positive.

\section{Results with Test-Time Filtering}
\label{sec:appendix-filters}

Table~\ref{tab:results-filters} reports, for each training configuration, the results obtained with the test-time filter that produced the highest \fbin. Since the filter was selected on the evaluation data and differs across rows, these results are exploratory and are not used to compare configurations.

\begin{table}[h]
\caption{Exploratory results with the best test-time filter for each training configuration, selected by \fbin on the evaluation data.}
\label{tab:results-filters}
\centering
\resizebox{\columnwidth}{!}{%
\begin{tabular}{lcccl}
\toprule
\textbf{Training} & \textbf{WER} & \textbf{\fbin} & \textbf{\fmacro} & \textbf{Best test filter} \\
\midrule
\multicolumn{5}{l}{\textit{NURC-Sampa (held-out test split)}} \\
Unfiltered & 0.106 & 0.730 & 0.857 & HP 600~Hz \\
LP 3200~Hz & 0.103 & 0.731 & 0.858 & HP 400~Hz \\
HP 400~Hz & 0.106 & 0.730 & 0.857 & HP 600~Hz \\
HP 600~Hz & 0.106 & 0.731 & 0.858 & HP 400~Hz \\
Data aug. & 0.145 & 0.728 & 0.856 & LP 1600~Hz \\
\midrule
\multicolumn{5}{l}{\textit{MuPe-Diversidades}} \\
Unfiltered & 0.162 & 0.792 & 0.888 & LP 1600~Hz \\
LP 3200~Hz & 0.132 & 0.779 & 0.881 & HP 600~Hz \\
HP 400~Hz & 0.132 & 0.795 & 0.890 & HP 200~Hz \\
HP 600~Hz & 0.135 & 0.796 & 0.890 & LP 3200~Hz \\
Data aug. & 0.135 & 0.775 & 0.879 & No filter \\
\bottomrule
\end{tabular}%
}
\end{table}

\end{document}